\documentclass{osa-article}
\usepackage{subfigure}
\usepackage{epstopdf}
\usepackage{amsmath}
\usepackage{multirow}
\usepackage{tabularx}
\usepackage{color}
\usepackage{multirow}
\usepackage{cite}
\usepackage{threeparttable}  
\usepackage{hhline}

\journal{boe}

\articletype{Research Article}

\begin{document}

\title{Analysis of CNN-based remote-PPG to understand limitations and sensitivities}

\author{Qi Zhan,\authormark{1} Wenjin Wang,\authormark{2,3*} and Gerard de Haan\authormark{3}}

\address{\authormark{1}Department of Electrical and Information Engineering, Hunan University, China\\
\authormark{2}Remote Sensing Group, Philips Research, The Netherlands     \\
\authormark{3}Electronic Systems Group, Department of Electrical Engineering, Eindhoven University of Technology, The Netherlands     }

\email{\authormark{* wenjin.wang@philips.com} }



\begin{abstract}
Deep learning based on Convolutional Neural Network (CNN) has shown promising results in various vision-based applications, recently also in camera-based vital signs monitoring. The CNN-based Photoplethysmography (PPG) extraction has, so far, been focused on performance rather than understanding. In this paper, we try to answer four questions with experiments aiming at improving our understanding of this methodology as it gains popularity. We conclude that the network exploits the blood absorption variation to extract the physiological signals, and that the choice and parameters (phase, spectral content, etc.) of the reference-signal may be more critical than anticipated. The availability of multiple convolutional kernels is necessary for CNN to arrive at a flexible channel combination through the spatial operation, but may not provide the same motion-robustness as a multi-site measurement using knowledge-based PPG extraction. Finally, we conclude that the PPG-related prior knowledge is still helpful for the CNN-based PPG extraction. Consequently, we recommend further investigation of hybrid CNN-based methods to include prior knowledge in their design.
\end{abstract}

\section{Introduction}

Remote Photoplethysmography (remote-PPG) is a contactless way to measure human cardiovascular activity by measuring the reflection variations of the skin registered by a video camera~\cite{verkruysse2008remote}. Over the last decade, various remote-PPG methods~\cite{lewandowska2011measuring,poh2010advancements,de2013robust,wang2016algorithmic,wang2015novel,de2014improved} have been proposed for PPG-signal extraction. The methods differ in their choice of assumptions~\cite{lewandowska2011measuring,poh2010advancements,de2013robust,wang2016algorithmic} and use of handcrafted features (e.g. projected color features of CHROM~\cite{de2013robust} and POS~\cite{wang2016algorithmic}), while these choices affect their robustness with respect to illumination variations and subject motion.\par
\subsection{Related work}
Recently, the success of deep Convolutional Neural Network (CNN) methods that automatically learn relevant features from images/videos in various applications has inspired researchers to attempt CNN-based remote-PPG extraction\cite{chen2018deepphys,niu2018synrhythm,vspetlik2018visual,yu2019remote1,yu2019remote}. Chen and McDuff~\cite{chen2018deepphys} proposed a convolutional attention network consisting of two parallel models to extract the PPG signal from a video. The first model is a classical "appearance model"~\cite{tran2017two} that learns to find the skin region-of-interest (RoI), while the second parallel path fed with DC-normalized frame-differences from the RoI learns to extract the PPG signal, using a finger oximeter-derived signal as a reference. In~\cite{chen2018deepphys}, the second model is referred to as a "motion model", but we prefer to use the term "normalized frame difference model", since our work shall prove that it exploits the blood absorption variation rather than the skin motion as suggested by~\cite{chen2018deepphys}. SynRhythm~\cite{niu2018synrhythm} is a general-to-specific transfer learning method. The authors directly convert the spatial-temporal features into heart rate based on the pre-trained network\cite{russakovsky2015imagenet}. HR-CNN~\cite{vspetlik2018visual} consists of the extractor CNN and the HR-estimator CNN with different loss functions to predict the heart rate, rather than the PPG signal. PhysNet~\cite{yu2019remote1} is a 3D CNN which learns the temporal and spatial context features of face sequences simultaneously to extract the PPG signal and then measures the heart rate variability from the PPG signal. Yu~\cite{yu2019remote} proposed a two-stage network that first enhances the quality of compressed videos and then retrieves the PPG signal from the enhanced videos. The main contribution of~\cite{yu2019remote} is using CNN to enhance the compressed videos, which is not the focus of this paper.\par

\subsection{Goal of this paper }
The benefit of CNN shown in various applications is that it enables good results without the need for the designer to analyze the problem in depth. The result, however, is a black box, which apparently performs the task it has been trained for. Since it is unclear how these results have been achieved, it is hard to predict the limitations of the system and to guarantee a proper training. We consider the results from CNN-based remote-PPG extraction interesting, but the lack of understanding may prevent further improvements and prove problematic for future healthcare applications. Hence, our paper aims at trying to understand the essence that the network has learnt and find its limitations when designing such a system. We do so by posing four questions, and design four experiments to answer them correspondingly.\par

The first question we try to answer is whether CNN extracts the pulse from skin absorption variations due to blood volume changes (PPG origin~\cite{de2013robust,wang2016algorithmic,wang2015novel,de2014improved}), or from balistocardiographic motion caused by blood pulsation (BCG origin~\cite{balakrishnan2013detecting}), or a combination of both. The understanding of this origin is crucial, since only if the CNN learns blood volume changes, additional vital signs measurement~\cite{tarassenko2014non,guazzi2015non}, e.g. SpO$_2$, will be possible. Thus the first question is: \textit{Does CNN learn PPG, BCG, or a combination of both?} \par 

The second question we aim to answer is regarding the reference PPG sensor (finger oximeter) used for CNN training. The relevance, here, is that the camera PPG signal measured from the face has a physiological delay as compared to the reference PPG signal measured at the finger. The pulse transit time from finger to face leads to the phase-shift between the PPG signal extracted from the finger oximeter and camera. Thus the second question is: \textit{Can the finger oximeter be directly used as the reference for CNN training?}\par

The third question we intend to address is whether the appearance of the skin affects the performance of CNN. The relevance of this question is that conventional remote-PPG methods are often used to extract the signal from the face, but they also work on other body parts that show pulsatile information (e.g. palm) or a small fragment of the face. The principles of the existing remote-PPG methods are generic and not limited to skin appearance. If CNN learns the distribution of the PPG-strength over the the entire face (similar to a PPG imager that shows the spatial distribution of physiological information), it will be difficult to generalize to other body-parts that have not been trained. Thus the third question is: \textit{Does CNN learn the spatial context information of the measured skin?} \par

The final question concerns the expected motion robustness. We hypothesize that motion robustness has been achieved by the projection of the camera color signal onto a direction that is orthogonal to the motion-induced distortions (i.e. allowing maximally 2 distortions for 3 color channels)\cite{wang2017robust}. The coefficients that define this projection direction will depend on the current distortions. Such flexibility of the CNN (i.e. various projection directions defined by CNN are dependent on motion-induced distortions) is only possible if there are different paths towards the final output that can be combined adaptively. The final question, therefore, is: \textit{Is CNN robust to motion and how is this motion-robustness achieved?}\par

The remainder of this paper is structured as follows. In Section 2, we introduce the CNN method selected for PPG extraction. In Section 3, four experiments are designed to answer the proposed research questions. In Section 4, we discuss the experimental results. Finally, in Section 5, we draw our conclusions and introduce our future options.\par

\section{Methods}
\begin{figure}[!htb]
\centering\includegraphics[width=\textwidth]{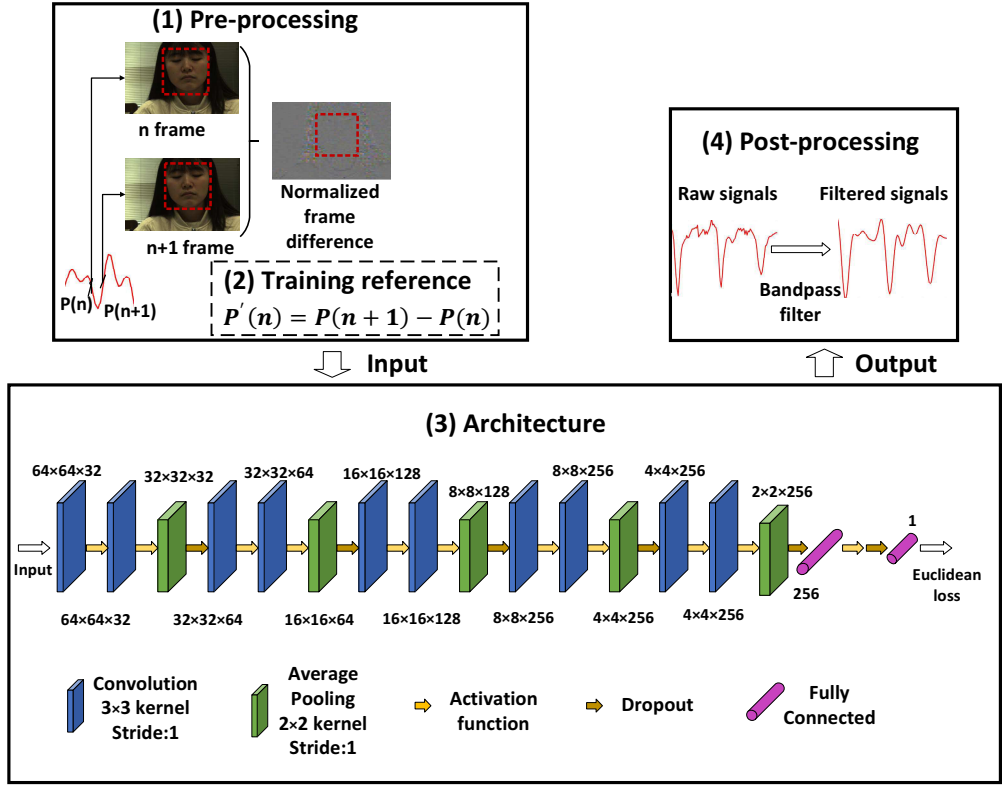}
\caption{The CNN-PPG flowchart includes four major parts: (1) Pre-processing: normalized image difference (AC/DC image); (2) Training reference: the derivative of the reference PPG signal; (3) The architecture of the CNN used in our experiments (ten convolution layers, five pooling layers, two fully-connected layers); (4) Post-processing: a bandpass filter with low cut-off frequency of 0.7 Hz and high cut-off frequency of 3 Hz.}
\label{fig: architecture}
\end{figure}

In this section, we introduce the CNN model used for PPG measurement and its settings. The workflow of the CNN is shown in Fig. \ref{fig: architecture}, and the components (e.g. input data, training label, and architecture) are introduced in the following subsections.

\subsection{The CNN training and testing input}
Since we adopt the CNN model of~\cite{chen2018deepphys} for our investigation, the input of CNN (both training and testing) remains the same as~\cite{chen2018deepphys}, i.e. the DC-normalized pixel difference between two consecutive frames in an image sequence (see Fig. \ref{fig: architecture}). The reason why the temporally normalized image difference (AC/DC image) is used has been explained in~\cite{wang2014exploiting}. The AC/DC is defined as the amplitude variation of the target signal after temporal normalization (i.e. dividing the signal by its DC). The AC/DC eliminates the dependency on the DC color of skin and illumination intensity, thus allowing the CNN to learn the color variation characteristics of the skin reflection.

\subsection{The CNN training label}
The label used for CNN training is the differentiated PPG signal that corresponds to the input of temporally normalized image difference. Both represent the derivative information. In this work, we use the PPG signal extracted by POS~\cite{wang2016algorithmic} as the reference label for training rather than the PPG signal measured by the finger oximeter. The reason is that the POS-signal is measured from the same video data (e.g. skin pixels of a face), which does not have any physiological delay (due to pulse transit time), as the PPG signal measured at the finger. The training label and video data are therefore fully aligned in time. We mention that the reason of choosing POS for creating training label set is that it shows general robustness under different circumstances and its performance has been reproduced by various works~\cite{vspetlik2018visual,wang2017robust,wu2017camera,gudi2019efficient}, although other motion-robust methods could be used equally.

\subsection{The CNN architecture}
The CNN architecture is shown in Fig. \ref{fig: architecture}. We create a "normalized frame difference model" that is similar to~\cite{chen2018deepphys} to learn the mapping between temporally normalized image difference and training label.  Since the focus of this paper is to understand the principle of CNN-based PPG signal extraction, we do not use the "attention model" used by~\cite{chen2018deepphys} that plays as an auxiliary role in finding the skin RoI (e.g. face). The "attention model" is useful for improving the robustness of measurement, but not essential for us to answer the questions regarding the PPG extraction. Therefore, it is eliminated for our investigation, in order to objectively show the performance of CNN-based PPG signal extraction, but not CNN-based skin detection. To ensure that the CNN learns the skin pixel color changes, we manually crop the face area and use it as the input.\par

Our CNN architecture contains ten convolution layers. The size of the used convolution kernel is $3\times3$. After each two convolution layers, an average pooling layer with $2\times2$ kernel is inserted, followed by a dropout layer to prevent overfitting. The activation function is hyperbolic tangent (tanh), which follows the convolution layer. Similar to other CNN regression tasks~\cite{chen2018deepphys}, the Euclidean distance is used as the loss function to measure and minimize the difference between the fully-connected layer and label (i.e. differentiated PPG signal). During the training process, the size of the temporally normalized image is $64\times64$ (i.e. with only the subject face), and the batch size is set to 32. The Adadelta is used and its optimizer parameters are set according to~\cite{zeiler2012adadelta}. \par

\section{Experiments}
\subsection{Video dataset}
$\bullet$ \textbf{HNU dataset}. The HNU dataset is a private dataset created by Hunan University for synthetic experiments in understanding the CNN-based remote-PPG measurement. It is a clean dataset without practical challenges such as body motions, which allows us to investigate the network deterministically. It contains a total of 26 video sequences recorded on 26 participants (19 males and 7 females, aged between 23 to 26 years old) with skin-type III according to the Fitzpatrick scale. The study was approved by Hunan University and informed consent was obtained from each subject. During the recording, the subject sits still in front of the camera with his/her face captured by a regular RGB camera (Global shutter RGB CMOS camera USB M2ST036-H from Shenzhen city Shen Technology Co. Ltd.) for one minute duration. The video is saved in the lossless BMP format (480$\times$320 pixels, 8-bit depth) and constant frame rate at 20 frames per second (fps). \par

$\bullet$ \textbf{PURE dataset}. The Pulse Rate Detection Dataset (PURE)~\cite{stricker2014non} consists of the subjects (8 males and 2 females) performing six types of head motions (i.e. steady, talking, slow translation of head, fast translation of head, small rotation of head, and medium rotation of head) in front of the camera. In total 59 video sequences are captured by an eco274CVGE camera at 30 fps with a cropped image resolution of 640$\times$480 pixels and 8-bit depth (The dataset does not provide the video recording of subject 6 with taking). Each video is recorded in one minute duration and saved in the lossless PNG format. The ground truth is the contact-based PPG signal sampled by a finger pulse oximeter (pulox CMS50E) with a sampling rate at 60 Hz.  \par

\subsection{Evaluation metrics}
We use two metrics to evaluate the performance of CNN-based PPG extraction.

$\bullet$ \textbf{Root-Mean-Square Error} The Root-Mean-Square Error (RMSE) is used to measure the difference of pulse rates obtained by CNN and reference PPG sensor. A sliding window (256 samples window length, 1 sample sliding step) is used to compute the CNN-based PPG-rate trace and reference PPG-rate trace based on the input PPG signal. The PPG rate of each sliding window is estimated in the frequency domain by using the index of the maximum frequency peak between [40, 240] beats per minute (bpm). The PPG rates estimated in the sliding window are concatenated into a full video trace for RMSE analysis. The RMSE represents the sample standard deviation of the absolute difference between reference and measurement.\par

$\bullet$ \textbf{Accuracy} It refers to the percentage of video frames where the absolute difference between the CNN-based PPG-rate and reference PPG-rate is smaller than 3 bpm.\par

\subsection{Benchmark protocols}

To answer the four questions we proposed in the introduction, we design four experiments accordingly, described as follows. The HNU dataset is divided into six groups, four videos per group. The exact split of groups is shown in Table \ref{table1}. All the below experiments are trained and tested on five-fold cross validation. All experiments are implemented on lossless encoded data.
\begin{table}
\begin{center}
\caption{The exact split of groups in the HNU dataset.}

\label{table1}

\begin{tabular}{|c|  c|  c|  c|  c|  c | c|} 
\hline

Group numbers  & 1 & 2 &  3 &  4 &  5 & 6\\
\hline
Subject numbers  & 1-4 & 5-8 & 9-12 & 13-16  &  17-20 & 21-24\\
\hline
\end{tabular}
\end{center}
\end{table}

$\bullet$ \textbf{Experiment on the measurement origin of CNN}
This experiment aims at answering the first question of whether CNN extracts the PPG, BCG, or a combination of both. Since the PPG source (i.e. blood absorption variation) is wavelength dependent, different wavelength channels of an RGB camera contain different relative pulsatile strength (AC/DC), i.e. the G channel has the strongest pulsatility, followed by the B channel and R channel. This color characteristic has been exploited by the knowledge-based algorithms for PPG extraction~\cite{de2013robust,wang2016algorithmic}. In contrast, the BCG source (i.e. head motion) is not wavelength dependent. To verify whether CNN learns the PPG, we use the channel order of R-G-B for CNN training and the channel order of R-B-G and B-G-R for CNN testing. If CNN learns the PPG, the changed channel order will fail to extract the PPG signal. If CNN learns the BCG, the channel order change will not influence the extraction. For further proof, we train and test the CNN using the video data with only the R channels (R-R-R), G channels (G-G-G) and B channels (B-B-B), separately. Since the G and R channels have strong pulsatile amplitude contrast but not the BCG, the CNN trained on the G-G-G will not differ from that trained on the R-R-R or B-B-B if it only learns the BCG motion.\par

$\bullet$ \textbf{Experiment on the training reference of CNN}
This experiment is to understand whether the physiological delay (due to pulse transit time) between the reference finger-PPG signal and the camera PPG signal measured from face is critical for CNN training. To this end, we design a synthetic experiment on the HNU dataset and a realistic experiment on the PURE dataset, respectively. For the first experiment, we deterministically simulate different phase delays between the reference PPG signal and camera PPG signal for training. Specifically, we define 6 phase delays ranging from 0 degree to 180 degrees with an interval of 30 degrees distance. Since the POS-signal is measured from the skin pixels of a face, it does not have any physiological delay as the PPG signal measured at the finger. Therefore, we can ensure that there is no phase shift between image color changes and training labels. Specifically, we use the POS-signal as the reference PPG signal (without phase delay). Then we modify the phase of the POS-signal using the Hilbert transform to create 5 different reference signals. Hence, the CNN has been trained on 6 different reference labels (with different phases w.r.t. the video content) given the same video input. \par

For the second experiment, we train the PURE dataset on four kinds of reference labels to investigate the actual influence of physiological delay caused by the finger oximeter. The subject 1 to subject 5 in the PURE dataset are used as training data and the rest subjects are used as test data. The ground truth of the PURE dataset is the contact-based PPG signal sampled by a finger oximeter. So we train the PURE dataset on the ground truth (called Finger-PPG) as a default setting. For the second test, we use the camera-PPG signal extracted by POS as the reference label to train the CNN (called Camera-PPG, no physiological delay). For the third test, we eliminate the phase shift between the finger-PPG signal and camera-PPG signal to align the finger-PPG signal with the video using the Hilbert transform (called Phase corrected finger-PPG). And we use the phase corrected finger-PPG signal as the reference label to train the CNN. Since the measurement principles of finger-PPG signals (transmission) and camera-PPG signal extraction (reflection) is different, the morphology of the camera-PPG signal and finger-PPG signal could be different, i.e. the finger-PPG signal has more high-frequency harmonics than the camera-PPG signal. Therefore, for the last test, a bandpass filter with low cut-off frequency of 0.7 Hz and high cut-off frequency of 3 Hz is applied to the phase corrected finger-PPG signal to eliminate the high-frequency harmonics (called Phase corrected and filtered finger-PPG). And we train the CNN using the phase corrected and the filtered finger-PPG signal as the reference label to explore whether the difference of waveform will influence the CNN training.\par

$\bullet$ \textbf{Experiment on the spatial context learning of CNN}
Since the appearance features are eliminated when using the normalized difference image (AC/DC image) as the input for CNN training and testing, we doubt whether the operation of spatial convolution that aims at combining different spatial context is still essential for the network. Thus we test different image resolutions (with different level of details of the appearance features). By default, the input image resolution for CNN training and testing is \(64\times64\) pixels, i.e. requirement of the network. We first down-scaled the input images from \(64\times 64\) to \(1\times1\), \(4\times4\), \(8\times8\), \(20\times20\), \(30\times30\), \(40\times40\), \(50\times50\) pixels, and then up-scaled the down-scaled images back to \(64\times64\) pixels as required by CNN. The images are down-scaled and up-scaled by the nearest-neighbor interpolation. We stress that although the input image resolutions are eventually the same, their appearance features are very different. Moreover, we rotate the test image by 90 degrees and 180 degrees, respectively. We test the CNN trained on images without rotation on these rotated images to see whether such dramatic changes on appearance will influence the extraction. \par
\begin{figure*}
\centering
\subfigure[\(1\times1\) ]{
\begin{minipage}[b]{0.12\linewidth}
\includegraphics[width=1\linewidth]{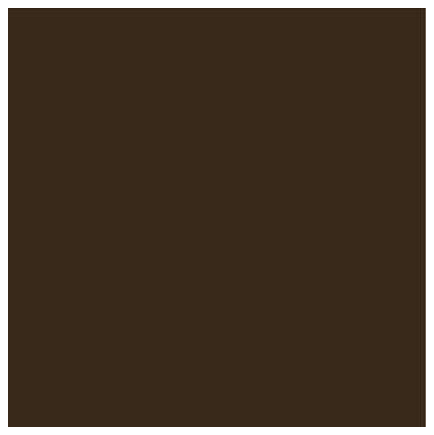}
\end{minipage}}
\subfigure[\(4\times4\) ]{
\begin{minipage}[b]{0.12\linewidth}
\includegraphics[width=1\linewidth]{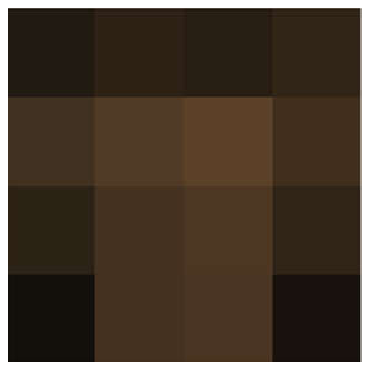}
\end{minipage}}
\subfigure[\(8\times8\) ]{
\begin{minipage}[b]{0.12\linewidth}
\includegraphics[width=1\linewidth]{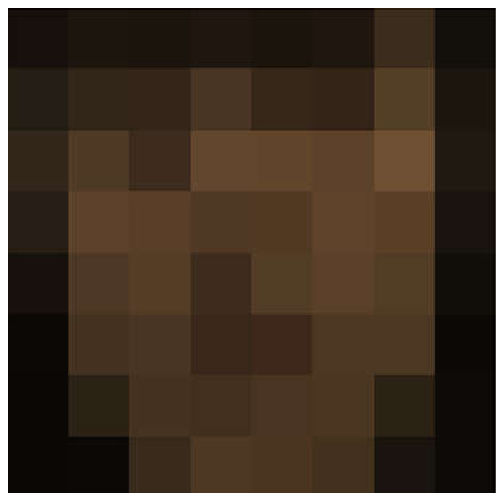}
\end{minipage}}
\subfigure[\(20\times20\) ]{
\begin{minipage}[b]{0.12\linewidth}
\includegraphics[width=1\linewidth]{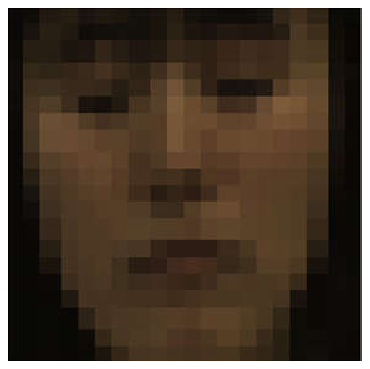}
\end{minipage}}
\subfigure[\(30\times30\) ]{
\begin{minipage}[b]{0.12\linewidth}
\includegraphics[width=1\linewidth]{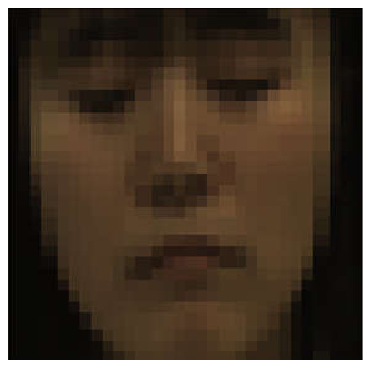}
\end{minipage}}
\subfigure[\(40\times40\) ]{
\begin{minipage}[b]{0.12\linewidth}
\includegraphics[width=1\linewidth]{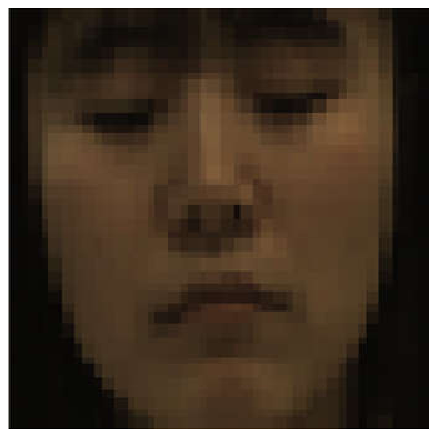}
\end{minipage}}
\subfigure[\(50\times50\) ]{
\begin{minipage}[b]{0.12\linewidth}
\includegraphics[width=1\linewidth]{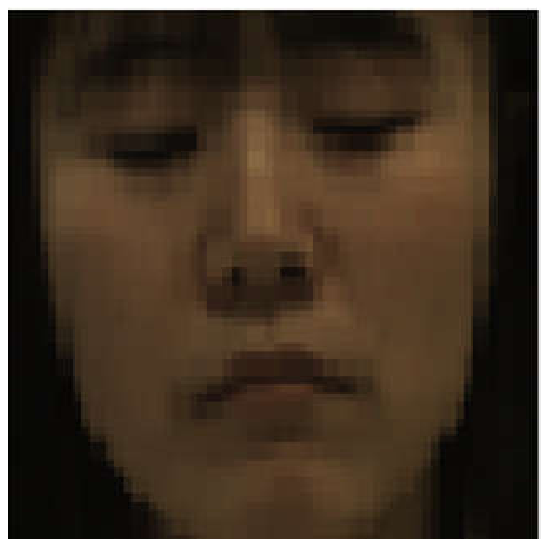}
\end{minipage}}
\caption{The snapshots of different image resolutions used in the experiment on the spatial context learning of CNN. (a)-(g) are the raw images (face area) with different image resolutions.}
\label{fig: resolution image}
\end{figure*}
$\bullet$ \textbf{Experiment on the motion robustness of CNN}
This experiment explores the motion-robustness of the CNN-based PPG signal extraction. A random Gaussian noise with the mean value $\mu$=0 and standard deviation $\sigma$=1 is added in each color channel of the HNU dataset to mimic noise disturbances. The noise AC-amplitude in the normalized R-G-B channels are the same for simulating the intensity variation on the temporally normalized skin-tone ($[1,1,1]$) direction\cite{wang2016algorithmic}. We use a mixture of clean video data (HNU dataset) and noise-perturbed video data for training (called CNN+Noise). Because it is difficult to identify whether the Gaussian noise is eliminated by CNN in the frequency domain, we add periodic noise perturbations in the test data for validation. The periodic noise in the test data has a peaked spectrum in the frequency domain. Thus, we can easily justify whether CNN has eliminated such noise. Besides, the use of different types of noise waveform in the test data can explore whether CNN learns to eliminate the noise or the Gaussian waveform. Therefore, a periodic noise (e.g. sinusoidal signal with the frequency of 1.67 Hz) is added in the R-G-B channels with the same gain. Additionally, we test the CNN (trained on the HNU dataset) on the PURE dataset as the cross validation to see whether CNN can generalize its performance to a different dataset made by different type of cameras in different conditions, i.e. the camera configuration (e.g. sampling rate) of the PURE dataset is different from that of the HNU dataset. \par

\section{Results and discussion}
In this section, we show the results of four experiments and analyze them.\par

\begin{figure*}
\centering
\begin{minipage}[b]{0.9\linewidth}
\centering
\includegraphics[width=1.0\linewidth]{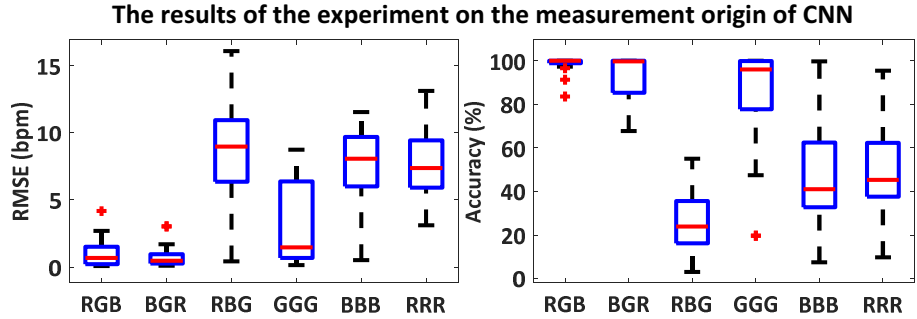}
\end{minipage}
\caption{The RMSE and accuracy of the CNN trained on the R-G-B channel order and tested on the R-G-B channel, B-G-R channel and R-B-G channel on the HNU dataset. And the RMSE and accuracy of the CNN trained and tested on the G-G-G channel, B-B-B channel and R-R-R channel order on the HNU dataset. In each panel, the median values are indicated by red bars inside the boxes, the quartile range by boxes, the full range by whiskers, the outliers by red cross. }

\label{fig: ex1}
\end{figure*}
$\bullet$ \textbf{Experiment on the measurement origin of CNN} Fig. \ref{fig: ex1} shows that the test data with the R-B-G channel order has larger RMSE than that with the R-G-B or B-G-R channel order. The accuracy of the test data with the R-B-G channel order (i.e. swap the G and B channel) is lower than 40\%. The RMSE obtained on the test data with the G-G-G setting is much smaller than that with the R-R-R setting. This is because the G channel contains the most pulsatile information (strong PPG component), whereas the R channel has much lower pulsatile amplitude. These results suggest that the PPG extraction of CNN is affected by the wavelength-channel order learnt during the training phase. It actually learns the wavelength-dependent characteristics of blood absorption variation and thus we expect that it measures the PPG signal, rather than the BCG-motion (i.e. motion-induced intensity changes) as suggested by the term "motion-model" in~\cite{chen2018deepphys}. 

\begin{figure*}
\centering
\subfigure[]{
\begin{minipage}[b]{0.9\linewidth}
\includegraphics[width=1.0\linewidth]{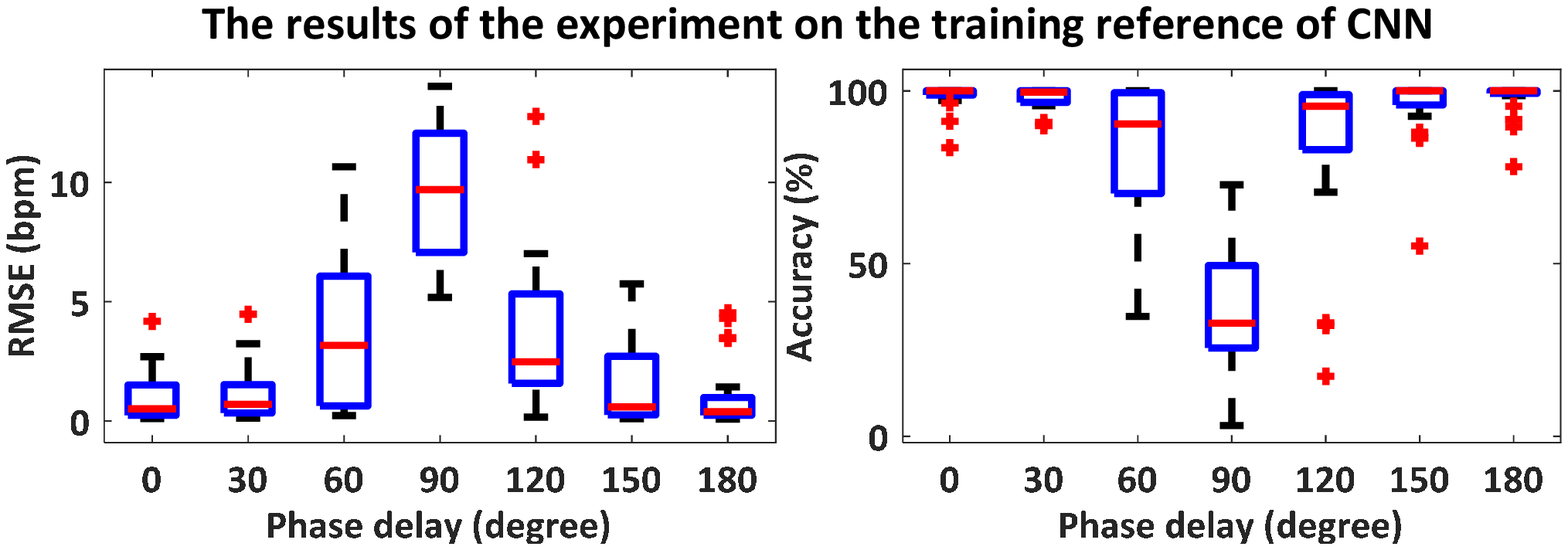}
\end{minipage}}
\subfigure[]{
\begin{minipage}[b]{0.9\linewidth}
\includegraphics[width=1.0\linewidth]{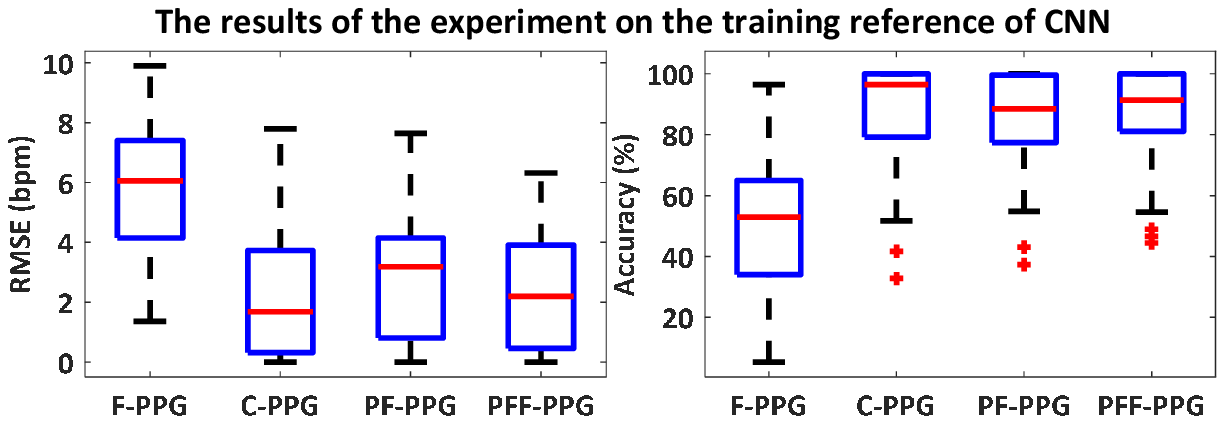}
\end{minipage}}
\caption{(a) The RMSE and accuracy of the CNN trained and tested on the HNU dataset by the labels with different phase delays; (b) The RMSE and accuracy of the CNN trained and tested on the PURE dataset with four kinds of labels. F-PPG: Finger-PPG; C-PPG: Camera-PPG; PF-PPG: Phase corrected finger-PPG; PFF-PPG: Phase corrected and filtered finger-PPG.}
\label{fig: ex2}
\end{figure*}

$\bullet$ \textbf{Experiment on the training reference of CNN} Fig. \ref{fig: ex2} (a) shows that the RMSE and accuracy vary with different phase delays. The RMSE increases with the phase increase of the trainning label till reaching the maximum at 90 degrees phase delay. After that, the RMSE decreases when the phase delay is changed from 90 degrees to 180 degrees. During the phase variation process, the accuracy is the smallest at 90 degrees phase delay. The RMSE and accuracy of the CNN trained on the label without phase delay are similar to that trained on the label with 180 degrees phase delay (with the opposite sign). The CNN test produces sign-flipped pulse signal but it does not cause a difference for pulse-rate calculation. Fig. \ref{fig: ex2} (b) shows that the Finger-PPG has the largest RMSE and smallest accuracy, suggesting that the physiological delay between the reference finger oximeter and video fails the CNN-based PPG extraction. The RMSE of the Phase corrected finger-PPG is smaller than that of the Finger-PPG. This is because the physiological delay between the reference finger oximeter and video data is eliminated. However, the RMSE of the Phase corrected finger-PPG is still larger than the Camera-PPG. We expect the reason to be the high-frequency harmonics of the phase corrected finger-PPG signal. When the high-frequency harmonics of the phase corrected finger-PPG signal are eliminated by the bandpass filter, the RMSE of the Phase corrected and filtered finger-PPG further decreases. Since there is no finger motion in the HNU dataset and PURE dataset, the phase corrected and filtered finger-PPG signal is more suitable than the camera PPG signal to be used as the reference to generate labels for CNN training. These results show that if there is a physiological delay between the reference finger oximeter and video data, the CNN training may not learn the correct match between the video content (e.g. skin color changes) and the reference PPG (i.e. inaccurate training due to signal phase shift). Moreover, the different morphology of PPG signal may influence the CNN training as well.

\begin{figure*}[!htb]
\centering
\subfigure[]{
\begin{minipage}[b]{0.9\linewidth}
\includegraphics[width=1.0\linewidth]{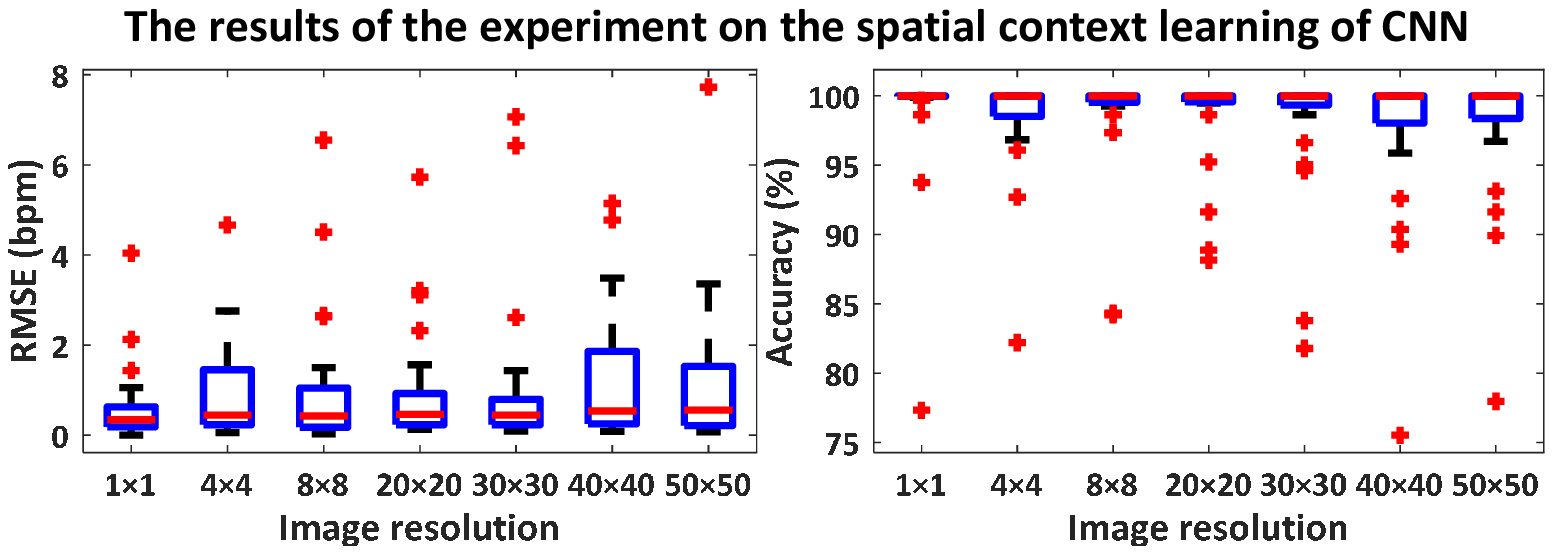}
\end{minipage}}
\subfigure[]{
\begin{minipage}[b]{0.9\linewidth}
\includegraphics[width=1.0\linewidth]{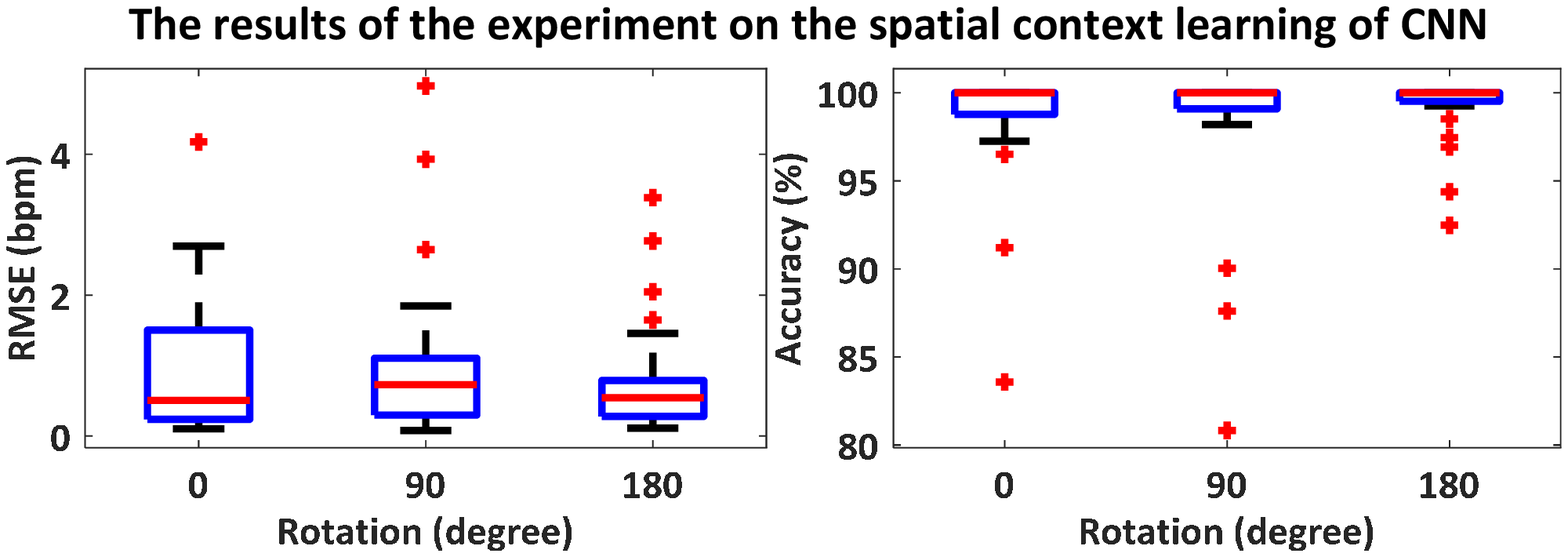}
\end{minipage}}
\caption{(a) The RMSE and accuracy of the CNN trained and tested on different image resolutions on the HNU dataset; (b) The RMSE and accuracy of the CNN trained on the HNU dataset and tested on the HNU dataset with different spatial context information (e.g. the images are rotated by 90 degrees and 180 degrees, respectively).}
\label{fig: ex3}
\end{figure*}

\begin{figure*}
\centering
\subfigure[RGB kernels]{
\begin{minipage}[b]{0.45\linewidth}
\includegraphics[width=1\linewidth]{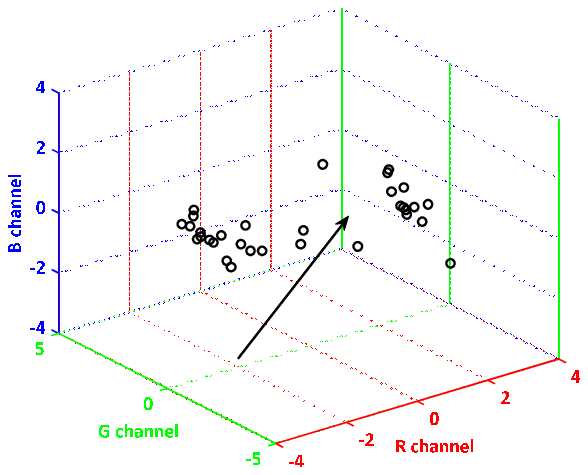}
\end{minipage}}
\subfigure[POS and CHROM kernels]{
\begin{minipage}[b]{0.45\linewidth}
\includegraphics[width=1\linewidth]{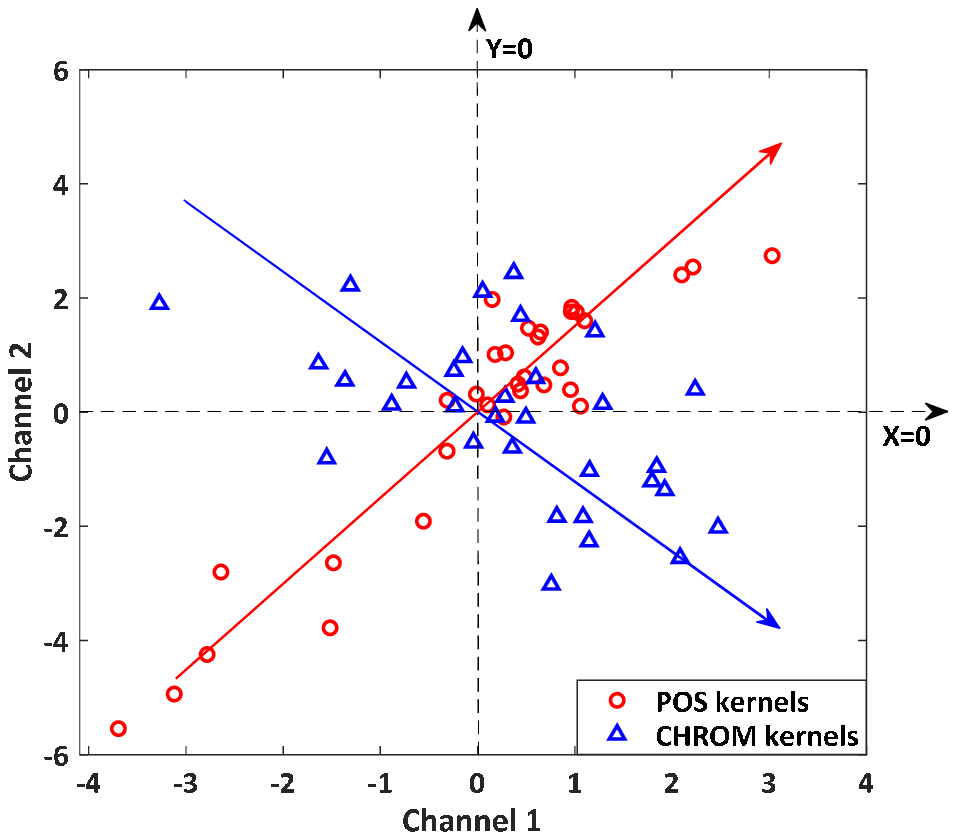}
\end{minipage}}
\caption{(a) The scatter plot of the sum of the weights in the R-G-B channels from the first convolution layer
trained on the image with 1$\times$1 pixel on the HNU dataset. Each scatter point represents the sum of the weights of each convolution kernel in the R-G-B channels. The black vector $[0.46,-0.85,0.25]^\mathrm{T}$ indicates the exact projection direction to combined the R, G and B channel. (b) The scatter plots of the sum of the weights in the 2 projected channels from the first convolution layer (CNN+POS, CNN+CHROM) trained on the HNU dataset. Each scatter point represents the sum of the weights of each convolution kernel in the two projected channels. The red vector $[0.55,0.83]^\mathrm{T}$ and the blue vector $[0.63,-0.78]^\mathrm{T}$  indicate the exact projection direction for the two projected channels in the CNN+POS and the CNN+CHROM, respectively. }
\label{fig: weight}
\end{figure*}

$\bullet$ \textbf{Experiment on the spatial context learning of CNN} Fig. \ref{fig: ex3} (a) shows that the different image resolutions cause small fluctuations in RMSE and accuracy. These results indicate that the CNN trained on different image resolutions can extract the PPG signal. When the CNN is trained on the images without rotation and is tested on the images with the rotation of 90 degrees and 180 degrees respectively, the variance of the RMSE is less than 0.3 bpm and the variance of the accuracy is less than 2\%. The results in Fig. \ref{fig: ex3} (b) show that the CNN can extract the pulse signal even if the appearance features of the test images are dramatically changed. We even find that the CNN trained on the frontal face can extract the PPG signal from other body parts (e.g. palm, side-view face). These results suggest that the appearance features are not important for CNN given the input of the normalized frame difference image (i.e. AC/DC pre-processing). 

The RMSE of the CNN trained on the resolution of \(1\times1\) pixel is smaller than other settings, which is a functional setting. It contains no appearance features. The convolution operation of CNN on a multi-channel image includes both the spatial operation and channel operation. Firstly, the spatial context of each channel is combined with the weights of the convolution kernel through the spatial operation. Then the results of the spatial operation on each channel are added to output the activation map. Since the initial weights of the convolution layer defined by the CNN has a uniform distribution, the weights of the convolution layer the CNN learnt from the training image with \(1\times1\) pixel do not necessarily have the same value. Although the image with \(1\times1\) pixel contains no appearance features (the appearance features of image are averaged into the \(1\times1\) pixel), the weights of the convolution kernels learnt by CNN are still diverse, and thus provide flexible paths (e.g. the projection direction shown in Fig. \ref{fig: weight} (a)) to combine the 3 color-channels into a pulse signal. This could also be understood as: if there is only one kernel, there is only one way of combining the R-G-B channels; if there are multiple kernels while kernels have different weights, there are multiple ways to combine the R-G-B channels. Therefore, the availability of multiple kernels/channels in the skin region is necessary to provide flexible paths (or freedom) for generating a pulse signal.\par

\begin{figure*}
\centering
\subfigure[Raw image]{
\begin{minipage}[b]{0.3\linewidth}
\includegraphics[width=1.0\linewidth]{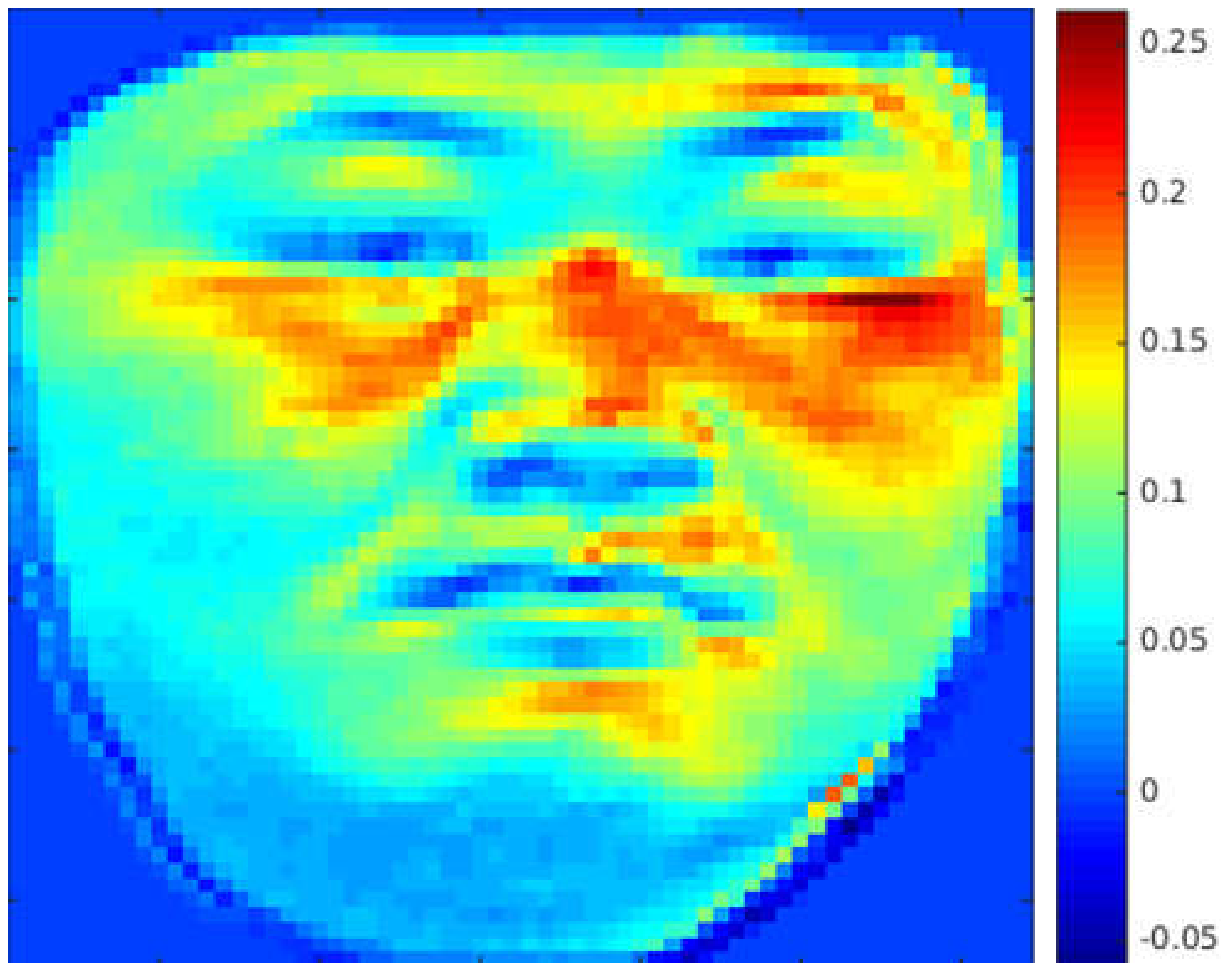}
\end{minipage}}
\subfigure[Normalized image difference]{
\begin{minipage}[b]{0.3\linewidth}
\includegraphics[width=1.0\linewidth]{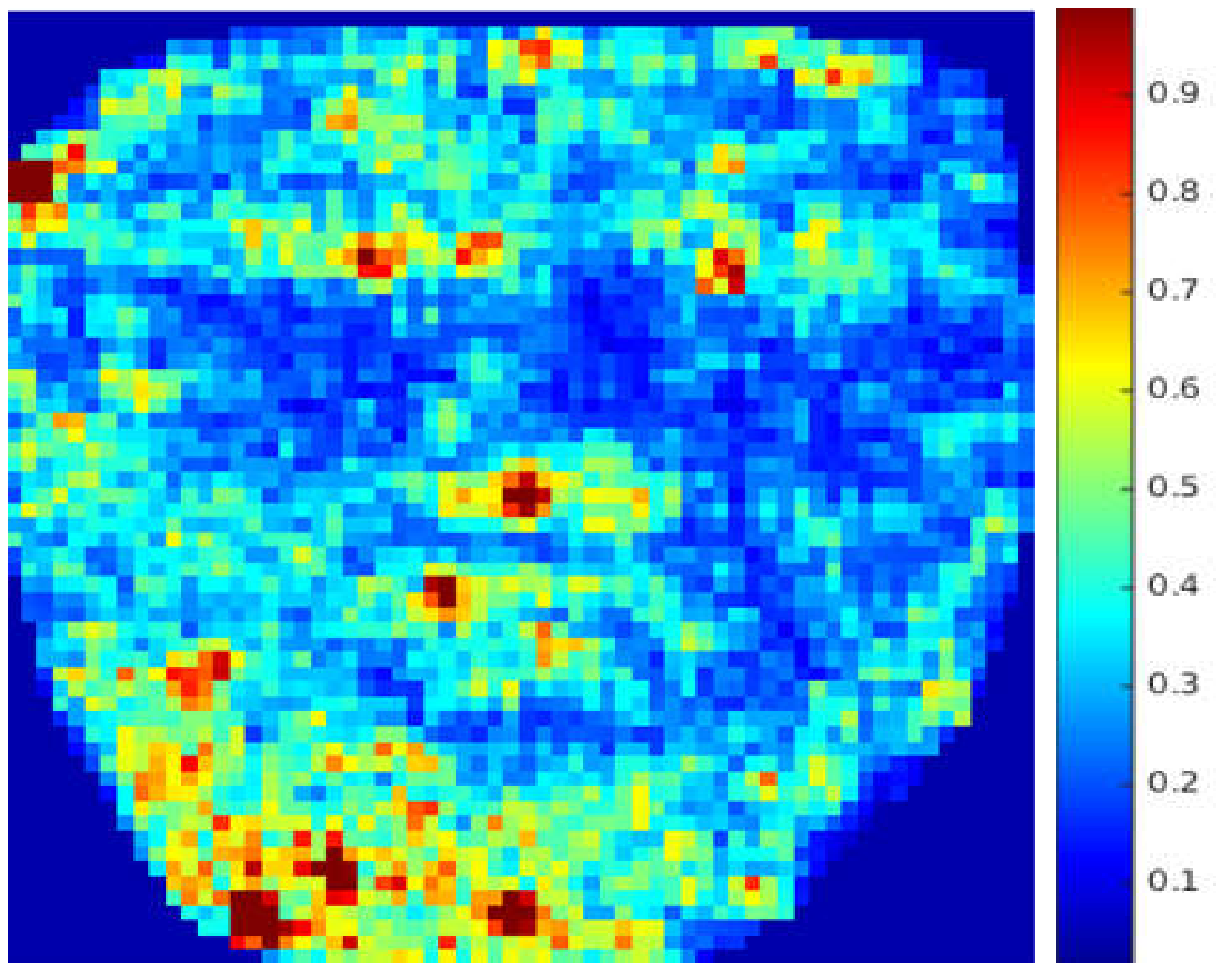}
\end{minipage}}
\subfigure[Palm]{
\begin{minipage}[b]{0.3\linewidth}
\includegraphics[width=1.0\linewidth]{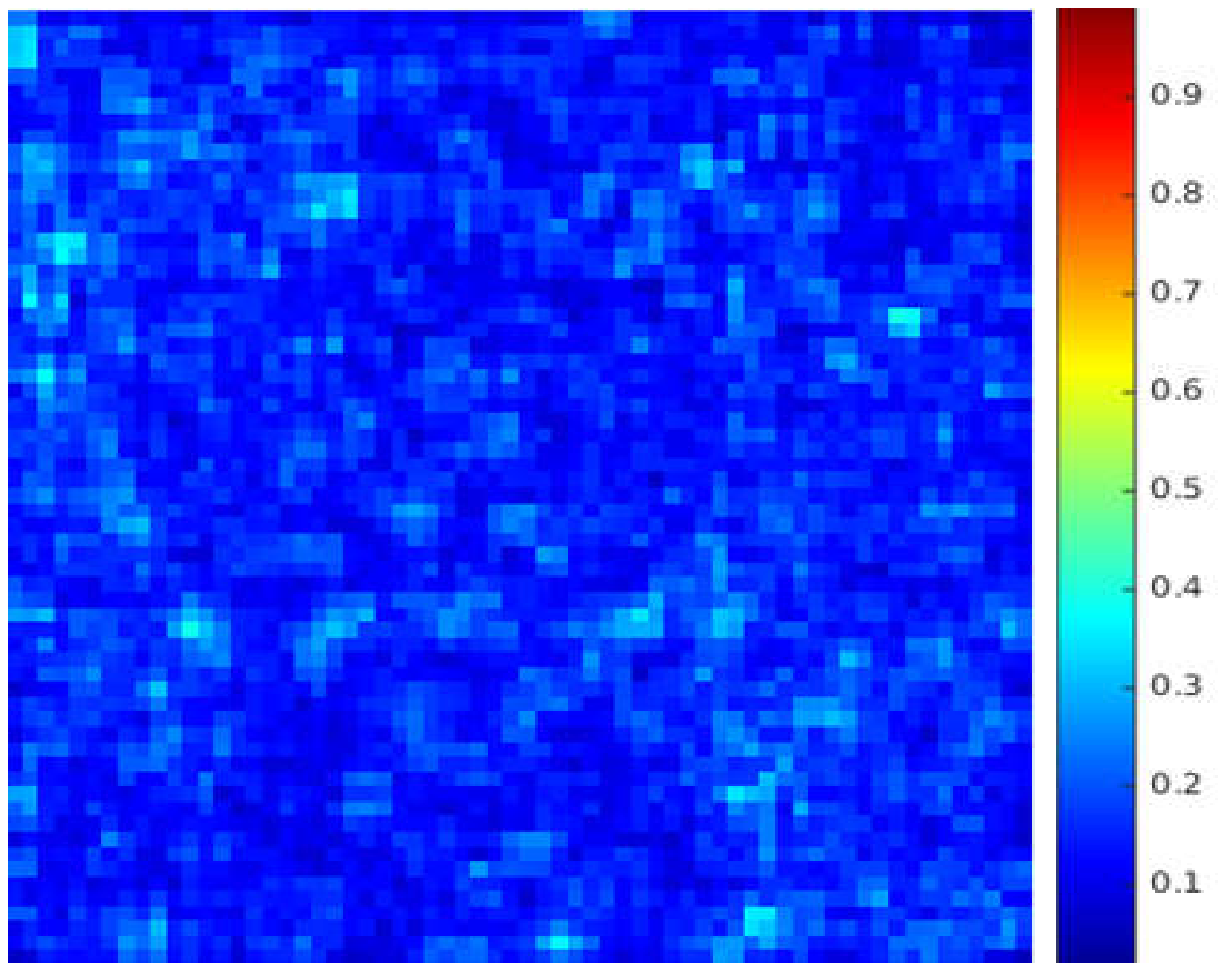}
\end{minipage}}
\caption{Visualization of the different activation maps extracted by the first convolution layer of the CNN trained on the HNU dataset: (a) The activation map extracted by the CNN from the raw DC image; (b) The activation map extracted by the CNN from the normalized frame difference image; (c) The activation map extracted by the CNN from the normalized frame difference image with the content of a palm. The bright areas indicate the larger weights in the activation map. }
\label{fig: featuremap}
\end{figure*}

Fig. \ref{fig: featuremap} shows the spatial weights resulting from CNN trained on a particular video from the HNU dataset. When the input image of CNN is the raw video frame with DC information (Fig. \ref{fig: featuremap} (a)), the high values in the activation map indicate that CNN focuses on the cheeks to extract the PPG signal. But the spatial weights resulting from CNN using the input of normalized frame differences (Fig. \ref{fig: featuremap} (b)-(c)) do not show a clear correlation with appearance features. This is because the facial features have been eliminated when the input of CNN is the normalized frame difference image. This input only shows the time-variations of the reflected light, which is not clearly correlated with the appearance features. These results suggest that appearance features are not available for the use of normalized frame difference image as CNN input. Therefore, the powerful operation of spatial convolution of CNN that aims at combining spatial contextual features is not achieved. But the different channel combination weights from different spatial locations enable the freedom of local projection.

Since the appearance features are not critical for CNN given the input of DC-normalized frame difference,  the training of CNN on different objects (e.g. face or palm) does not make a major difference. This confirms our expectation that the spatial redundancy of the PPG-measurements is not exploited by CNN (e.g. CNN cannot focus on the forehead of a talking subject which has stable PPG-strength than the mouth region, because the appearance features have been eliminated by the normalized frame difference image), but rather allows flexible combination of pixels with different color-channel combination weights by convolution operation. In knowledge-based PPG-extraction methods\cite{de2013robust,wang2016algorithmic,de2014improved}, this flexibility is hand-designed and the performance dose not depend on the training data. 
 \par
\begin{figure*}[!htb]
\centering
\begin{minipage}[b]{0.9\linewidth}
\includegraphics[width=1.0\linewidth]{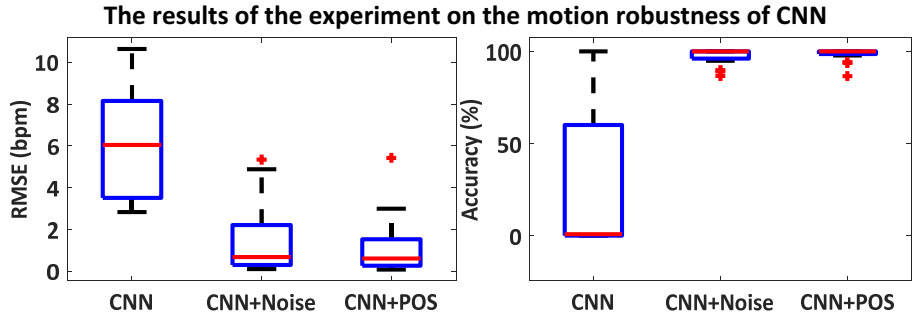}
\end{minipage}
\caption{The RMSE and accuracy of the CNN, CNN+Noise and CNN+POS trained on the HNU dataset and tested on the noisy-perturbed (periodic noise) video data.}
\label{fig: ex4}
\end{figure*}

$\bullet$ \textbf{Experiment on the motion robustness of CNN} Fig. \ref{fig: ex4} shows that the RMSE of the CNN trained on clean videos (HNU dataset) and tested on noisy videos is larger than 5 bpm and the accuracy is close to 0\%. It suggests that the convolution kernels of CNN learnt from the clean videos cannot offer flexible projections to the motion-induced distortions present in the noisy videos. However, the RMSE of the CNN+Noise is smaller and the accuracy is close to 100\%. This is because the CNN+Noise is trained on a mixed data which contains both the clean video data and noise-perturbed video data. It can differentiate between the intensity variations caused by blood absorption and motion. \par

Inspired by the knowledge-based prior art POS that projects a three-channel (R-G-B) image onto a plane orthogonal to the $[1,1,1]$ direction to create a two-channel image for PPG extraction~\cite{wang2016algorithmic}, we use this prior knowledge for CNN to see whether it improves the performance of a pure CNN. Before feeding into the CNN, the normalized image differences from the clean videos are first projected onto a plane defined by POS (called CNN+POS). As the inputs are two-channel images, we change the number of channels (three channels) of the first convolution layer in the CNN into two channels. In Fig. \ref{fig: ex4}, the RMSE of CNN+POS is smaller than CNN and CNN+Noise. This is because that the synthetic periodic noise is along the intensity variation direction $[1,1,1]$ in the DC-normalized color space. POS projection can perfectly eliminate such distortions.\par

\begin{figure*}[!htb]
\centering
\subfigure[]{
\begin{minipage}[b]{0.9\linewidth}
\includegraphics[width=1.0\linewidth]{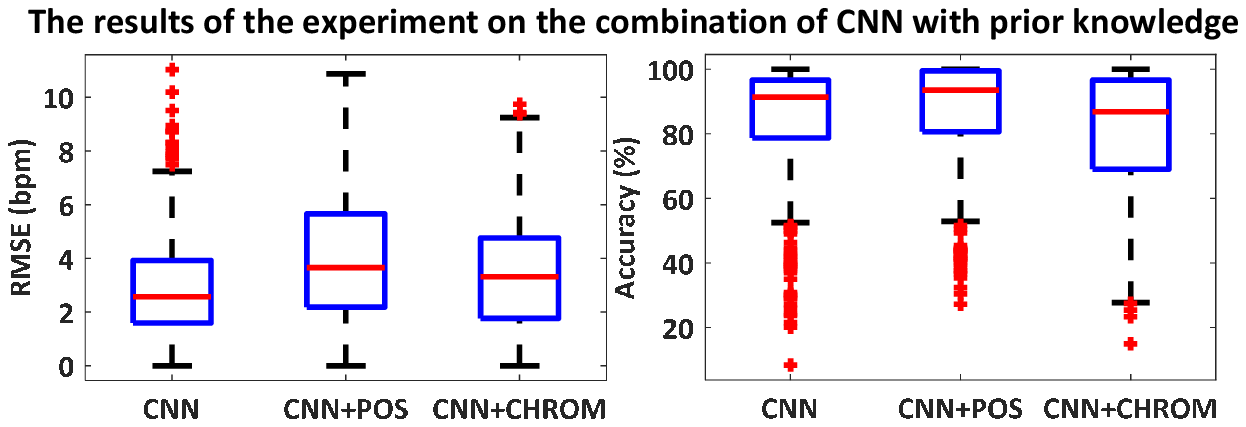}
\end{minipage}}
\subfigure[]{
\begin{minipage}[b]{0.9\linewidth}
\includegraphics[width=1.0\linewidth]{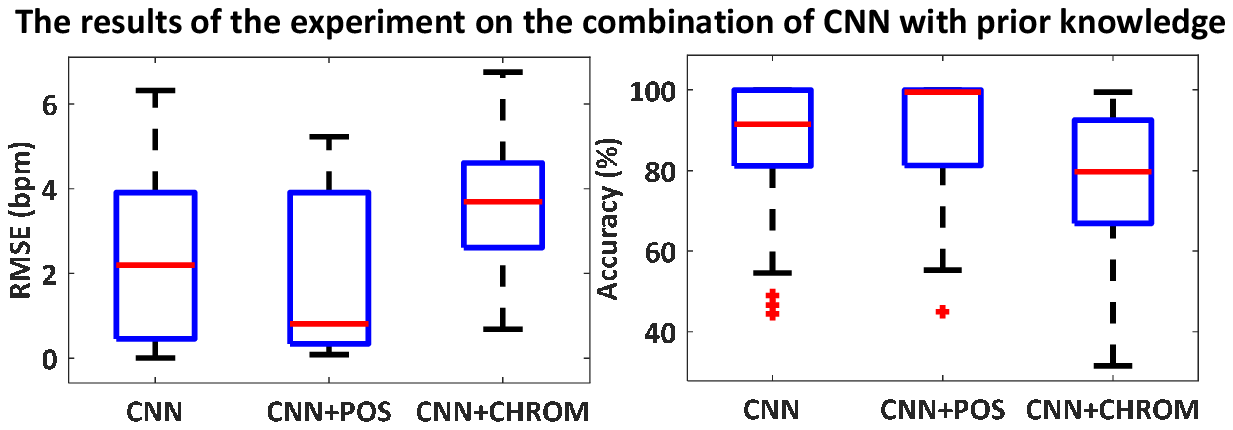}
\end{minipage}}
\caption{(a) The RMSE and accuracy of the CNN, CNN+POS and CNN+CHROM trained on the HNU dataset and tested on the PURE dataset; (b) The RMSE and accuracy of the CNN, CNN+POS and CNN+CHROM trained and tested on the PURE dataset.}
\label{fig: cross exI}
\end{figure*}
Since the two projection axes defined by POS give two inphase projected signals\cite{wang2016algorithmic}, POS adds the two inphase signals to boost the pulsatility of two projected signals. If the CNN+POS can eliminate the intensity variation noise  (as shown in Fig. \ref{fig: ex4}), it should learn the similar combination of the two projected signals defined by POS. To validate this hypothesis, we show the distribution of the convolution kernels in the first convolution layer of the CNN+POS in Fig. \ref{fig: weight} (b). It shows that the CNN+POS adds the two projected channels on the $[0.55,0.83]$ direction. According to the POS definition, the projected signals on the POS-plane show in-phase PPG components. Such a characteristic relationship has been learned by the CNN-POS kernel weights. An alternative approach of POS is CHROM, which uses two projection axes that are different from POS. The two projected CHROM-signals are anti-phase\cite{de2013robust}, thus CHROM subtracts the two anti-phase signals to strengthen the pulsatile components. In this sense, CNN+CHROM should also learn the characteristic channel combination of CHROM. Therefore, we project the normalized image difference from clean videos onto the CHROM-plane (orthogonal to the specular reflection direction)\cite{de2013robust} and train the CHROM-projection images. Fig. \ref{fig: weight} (b) shows that the two projected channels are combined on the $[0.63,-0.78]$ direction. The CNN-CHROM kernels weights are also in line with the CHROM definition, where the projected signals on the CHROM-plane show anti-phase PPG components. These results suggest that the prior knowledge-based CNN indeed learns the internal relationship between the projected channels defined by POS and CHROM and such relationship is associated with the relative PPG strength between the projected channels. This experiment further confirms the conclusion that CNN indeed learns the PPG origin.\par

 \begin{figure*}[!htb]
\centering
\subfigure[]{
\begin{minipage}[b]{0.9\linewidth}
\includegraphics[width=1.0\linewidth]{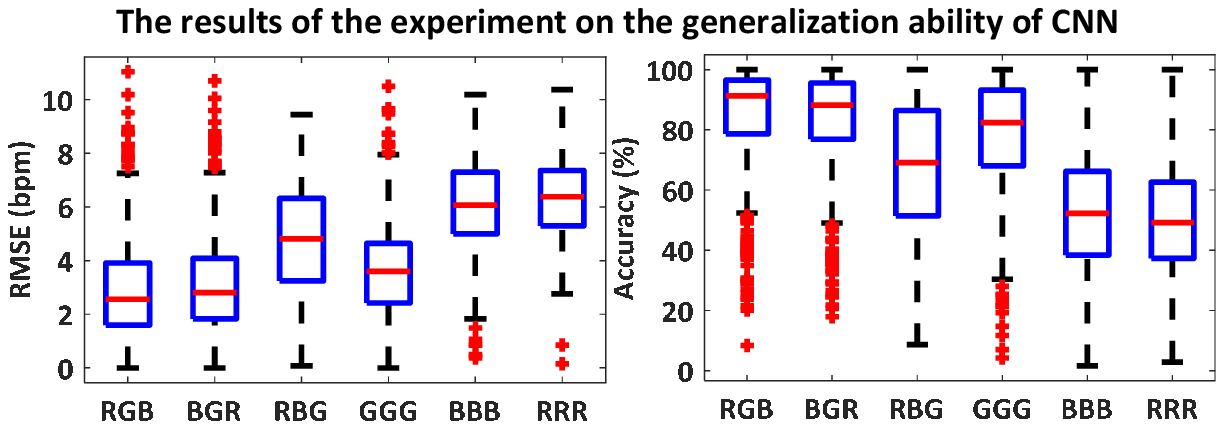}
\end{minipage}}
\subfigure[]{
\begin{minipage}[b]{0.9\linewidth}
\includegraphics[width=1.0\linewidth]{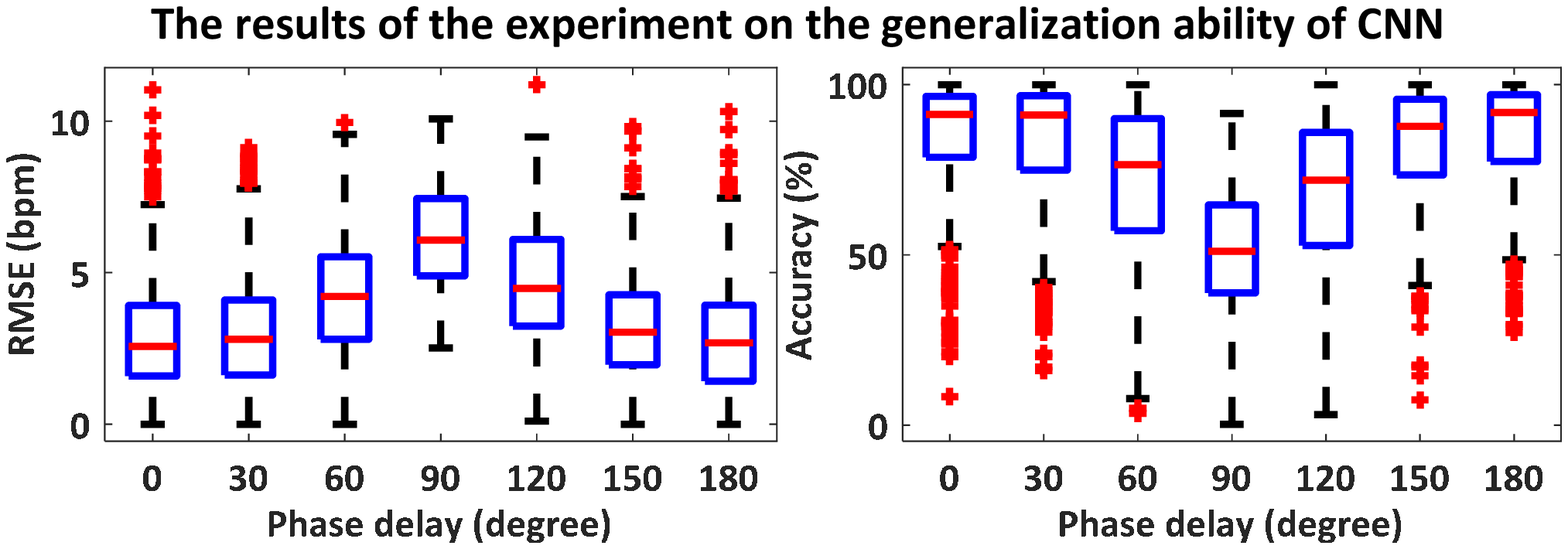}
\end{minipage}}
\caption{ (a) The RMSE and accuracy of the CNN trained on the HNU dataset with the R-G-B channel order and tested on the PURE dataset with the R-G-B-channel, B-G-R channel and R-B-G channel order. And the RMSE and accuracy of the CNN trained on the HNU dataset and tested on PURE dataset with the G-G-G channel, B-B-B channel and R-R-R channel order. (b) The RMSE and accuracy of the CNN trained on the HNU dataset  by the labels with different phase delays and tested on the PURE dataset. }
\label{fig: cross exII}
\end{figure*}

 \begin{figure*}[!htb]
\centering
\subfigure[]{
\begin{minipage}[b]{0.9\linewidth}
\includegraphics[width=1.0\linewidth]{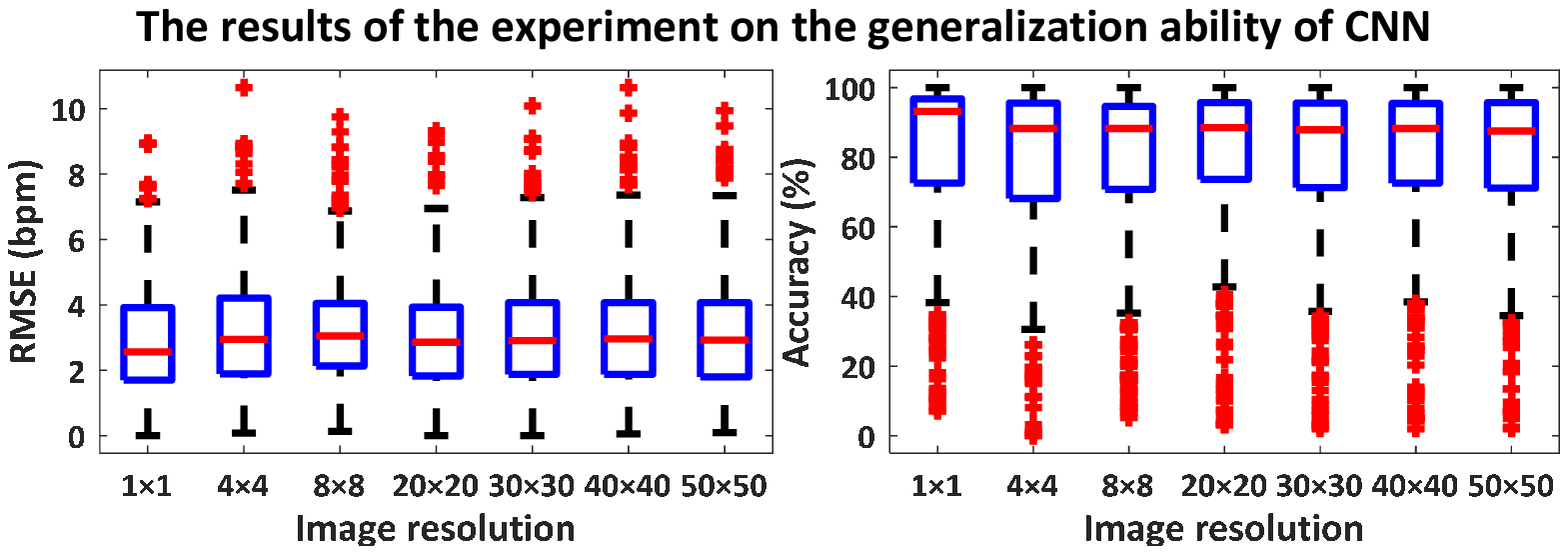}
\end{minipage}}
\subfigure[]{
\begin{minipage}[b]{0.9\linewidth}
\includegraphics[width=1.0\linewidth]{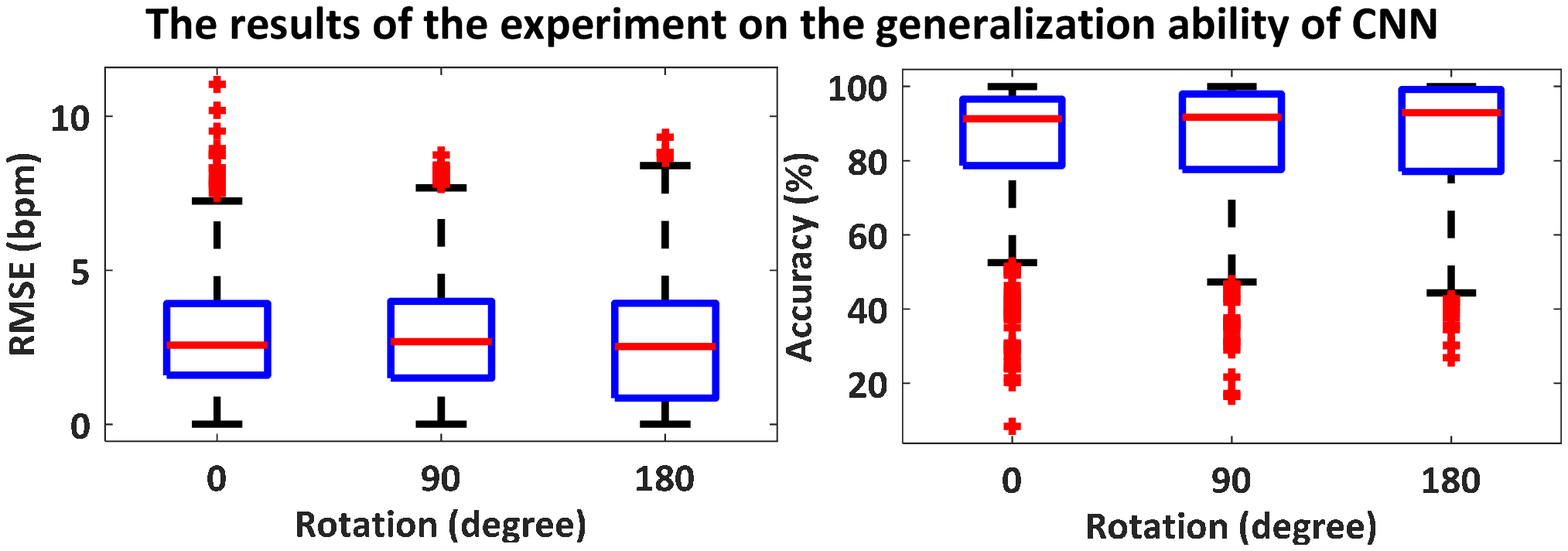}
\end{minipage}}
\caption{(a) The RMSE and accuracy of the CNN trained on the HNU dataset and tested on the PURE dataset  with different image resolutions; (b) The RMSE and accuracy of the CNN trained on the HNU dataset and tested on the PURE dataset with different spatial context information (e.g. the images are rotated by 90 degrees and 180 degrees, respectively). }
\label{fig: cross exIII}
\end{figure*}
In order to explore whether the prior knowledge-based CNN (CNN+POS, CNN+CHROM) is better than the pure CNN on realistic data, we test it on the challenging dataset (PURE dataset) based on two different training dataset. Firstly, we train the CNN, CNN+POS, CNN+CHROM on the clean dataset with stationary subject (HNU dataset) and test on the challenging dataset (PURE dataset) with six types of head motions. Secondly, the PURE dataset is divided into two groups and each group contains five subjects. The subject 1-5 are used for training and the subject 6-10 are used for testing. Fig. \ref{fig: cross exI} (a) shows that the CNN-based PPG extraction framework trained on the HNU dataset can be used for extracting the PPG signal from the PURE dataset, which suggests that CNN-based PPG extraction framework may be less sensitive to the parameters of recording (e.g. camera parameters, sampling rate). Besides, the RMSE of CNN+POS is smaller than CNN on the test videos with subjects' body motion (as shown in Fig. \ref{fig: cross exI}). However, the RMSE of CNN+CHROM is larger than CNN+POS and CNN. This is because that CHROM defines projection plane orthogonal to the specular distortion. When prior knowledge is inaccurate (i.e. the actual direction of the specular distortion is different from the direction of the specular distortion assumed in CHROM~\cite{de2013robust}), CNN+CHROM may perform worse than CNN and CNN+POS. These results show that the prior-knowledge may still improve CNN-PPG. Therefore, we recommend further investigation of hybrid CNN-based methods that include prior knowledge in their design.\par

 Fig. \ref{fig: cross exII} (a) shows the generalization ability of CNN. Although CNN is trained on the HNU dataset, it can still extract the PPG signal from the videos in the PURE dataset (i.e. cross dataset validation). In Fig. \ref{fig: cross exII} (a), the RMSE of CNN tested on the PURE dataset recording the fast translation of subject is the largest, which means the CNN (trained on clean videos) is sensitive to the larger-scale rigid motion. The larger RMSE and smaller accuracy of CNN tested on the R-B-G channel order demonstrate that CNN actually measures PPG. The variation of the RMSE of CNN trained on the reference with different phase delays validates the hypothesis that phase delays of the reference will affect the performance of CNN (as shown in Fig. \ref{fig: cross exII} (b)). Fig. \ref{fig: cross exIII} (a)-(b) show that the RMSE and accuracy of CNN are not altered by different image resolutions. \par

 These results demonstrate that CNN works (indeed learns the PPG but not only the BCG) and the performance of CNN could be reproduced (depends on dataset and CNN architecture), but its understanding needs to be improved, i.e. not only how it works but also why it works. The physiological delay of the PPG reference should be carefully considered when using the finger oximeter to acquire the training label. The CNN does work as it supposed to be (especially for the AC/DC image input), and the spatial contextual combination is much less important than channel combination. The flexible channel combination offers diverse projected planes for CNN to measure the PPG signal. \par

\section{Conclusions and Future Options}
In this paper, we used experiments to answer various questions to improve our understanding of CNN-based remote-PPG. From these experiments, we conclude that the network exploits blood absorption variation to extract the physiological signals rather than the balistocardiographic motion of the subject. We found that the choice and parameters (phase, spectral content, etc.) of the reference-signal may be more critical than anticipated. 

We could also conclude that the availability of multiple convolution kernels in the skin region is necessary for the method to arrive at a flexible color-channel combination through the spatial operation, but does not provide the same motion-robustness as a multi-site measurement using knowledge-based PPG extraction. This conclusion warrants future research comparing the CNN-PPG and knowledge-based methods in a "spatial redundancy framework", where both concepts extract the pulse signal locally and combine them to a global result. Since this is the state-of-the-art for knowledge-based methods, we would anyway consider it to be a more fair comparison. Besides, more experiments (with more practical challenges) would be performed to compare the CNN-PPG and the hybrid CNN-PPG that combines the prior knowledge of human physiology (e.g. POS or CHROM) with the CNN extraction. Furthermore, an experiment on a fully connected network with 3 input neurons would be interesting to investigate whether a fully connected network can achieve similar results as a CNN trained on the image resolution of of \(1\times1\) pixel. \par
Finally, we found that the prior knowledge still improves the results of the CNN in PPG extraction. Consequently, we recommend further investigation of hybrid CNN-based methods that include prior knowledge in their design. Since the CNN architecture used in this paper is based on DeepPhys, the conclusions derived in this study are restricted to this specific network. However, the approaches demonstrated in this paper in investigating a CNN-PPG method are generic and can be used to understand/diagnose other CNN-PPG methods. We expect the understanding/insights gained from this study to help the community to further improve the performance of CNN-based remote-PPG.\par

\section*{Funding}
This research is supported by the National Natural Science Foundation of China (Grant No. 61671204).

\section*{Acknowledgement}
The authors would like to thank Miss. Chuanxiang Tang for creating the HNU dataset and the volunteers from Hunan University for being test subjects.
\section*{Disclosures}
The authors declare that there are no conflicts of interest related to this article.

\bibliography{sample}

\end{document}